\documentclass[a4paper, 10 pt, conference]{ieeeconf}

\IEEEoverridecommandlockouts
\usepackage{graphicx}
\usepackage{subcaption}
\usepackage{amsmath} % assumes amsmath package installed
\usepackage{amssymb}  % assumes amsmath package installed
\usepackage{csquotes}
\usepackage{tabu}
\usepackage{balance}
\usepackage{tabularx}
\usepackage{array}
\newcolumntype{Y}{>{\centering\arraybackslash}X}
\usepackage{tikz}
\usetikzlibrary{calc, decorations.shapes, decorations.pathreplacing}

\newcommand{\hide}[1]{---HIDDEN---}
\renewcommand{\hide}[1]{#1} % for the camera ready version

\DeclareMathAlphabet{\mathcal}{OMS}{cmsy}{m}{n}
\DeclareMathOperator*{\argmin}{arg\,min}

\newcolumntype{C}[1]{>{\centering\let\newline\\\arraybackslash\hspace{0pt}}m{#1}}

\title{\LARGE \bf
Active and Transfer Learning of Grasps by Kernel Adaptive MCMC
}
\author{\hide{Philipp Zech, Hanchen Xiong and Justus Piater}%
\thanks{The research leading to these results has received funding from the \hide{European Community's Seventh Framework
        Programme FP7/2007-2013 (Specific Programme Cooperation, Theme 3, Information and Communication Technologies) under
        grant agreement no.~610532, SQUIRREL}.}%
\thanks{\hide{Philipp Zech, Hanchen Xiong, and Justus Piater are with the
        Institute of Computer Science,
        University of Innsbruck, 6020 Innsbruck, Austria.
        \texttt{philipp.zech@uibk.ac.at}, \texttt{hanchen.xiong@uibk.ac.at}, and \texttt{justus.piater@uibk.ac.at}}}%
}

\begin{document}

\maketitle
\thispagestyle{empty}
\pagestyle{empty}

%%%%%%%%%%%%%%%%%%%%%%%%%%%%%%%%%%%%%%%%%%%%%%%%%%%%%%%%%%%%%%%%%%%%%%%%%%%%%%%%%%%%%%%%%%%%%%%%%%%%%%%%%%%%%%%%%%%%%%%%%%%%%%%%%%%%%%%%%%%%%%%%%%%%%%%
\begin{abstract}
Human ability of both versatile grasping of given objects and grasping
of novel (as of yet unseen) objects is truly remarkable. This probably
arises from the experience infants gather by actively
playing around with diverse objects. Moreover, knowledge acquired
during this process is reused during learning of how to
grasp novel objects. We conjecture that this combined process of
active and transfer learning boils down to a random search around an
object, suitably biased by prior experience, to identify promising
grasps. In this paper we present an active learning method for learning
of grasps for given objects, and a transfer learning method
for learning of grasps for novel objects. Our learning methods
apply a kernel adaptive Metropolis-Hastings sampler that learns an
approximation of the grasps' probability density of an object while
drawing grasp proposals from it. The sampler employs simulated
annealing to search for globally-optimal grasps. Our empirical
results show promising applicability of our proposed learning schemes.
\end{abstract}
%%%%%%%%%%%%%%%%%%%%%%%%%%%%%%%%%%%%%%%%%%%%%%%%%%%%%%%%%%%%%%%%%%%%%%%%%%%%%%%%%%%%%%%%%%%%%%%%%%%%%%%%%%%%%%%%%%%%%%%%%%%%%%%%%%%%%%%%%%%%%%%%%%%%%%%

\section{Introduction}

Establishing efficient strategies for learning precision grasps is
one of the key challenges in robotics research. Currently, a substantial
amount of work in this area relies on an object's shape or shape-related
information (e.g., surface normals or image gradients) reconstructed
from vision~\cite{bohg2014} to analytically compute object specific
contact points~\cite{shimoga1996,bicchi2000}. A complementary idea though is
to neglect an object's shape and instead utilize its pose to learn feasible
grasps by sampling gripper poses relative to an object's pose.
This results in grasp learning methods that require very little additional
object specific knowledge (given its pose) to guide the search for feasible
grasps (cf.~Detry et al.~\cite{detry2011} who utilized 3D edge information).

Such a \emph{postural} interpretation of a grasp by a gripper's
pose relative to an object's pose possesses two key advantages compared
to shape-based grasp learning. First, learned grasps can be readily applied
to known objects by just aligning a gripper's pose relative to an object's
pose.  This requires no knowledge other than
an object's pose. Secondly, the postural interpretation  allows seamless
transfer of grasps between similar (e.g., in shape and size) objects by
just sampling a new gripper pose suitably biased by an already known gripper
pose. Conversely, a shape-based approach in both cases requires reconstructing
an object's shape which may easily fail due to clutter or improper segmentation.

Metropolis-Hastings (MH)~\cite{hastings1970} is a popular Markov-Chain Monte Carlo (MCMC)
algorithm that constructs a Markov chain on a state space $\mathcal{X}$ (e.g., the grasp parameter space)
where the stationary distribution of possible states is the target probability density
$\pi(x)$. By drawing samples $x_{0},x_{1},x_{2},\dots$ from a proposal distribution
$q(x|y$) one can iteratively approximate $\pi(x)$. We propose the application of
kernel adaptive MCMC (Section~\ref{sec:3}) for active learning of grasps for given
objects (Section~\ref{sec:4}) and transfer learning for acquiring grasps for novel
objects (Section~\ref{sec:5}) by learning, via approximation, an object's unknown grasp density.

In this paper we first introduce active learning of grasps for given objects by
MCMC Kameleon (Section~\ref{sec:4}). This relies on a rough sketch\label{sec:sketch}\footnote{Observe that
Sejdinovic et al.~characterize such a rough sketch as just a scheme with good exploratory
properties of the target; there is no need for it to result from a converged or even valid Markov chain.}
of the shape of the grasps' probability density $\pi$ of a specific object. Given this
rough sketch, MCMC Kameleon then learns an approximation to $\pi$ during its
burn-in phase for subsequent sampling of grasps. Secondly, we present transfer learning
to learn grasps for novel objects similar in \emph{shape and size} to objects already
learned (Section~\ref{sec:5}). This capitalizes on MCMC Kameleon's learning behavior
during its burn-in phase which allows learning grasps for different objects of the
same type (e.g., plates or soup plates) using a rough sketch of the shape and size
of a similar object.

A common problem in applying MCMC is deterioration of the sampler, i.e.,
the repeated sampling of rejected proposals. In our case, this amounts to
repeated sampling of gripper poses yielding infeasible grasps and thus zero success probability.
To tackle this problem, we utilize the object's rim computed from its point cloud
to define a meaningful probability measure for infeasible grasps. This allows us to heuristically
nudge the sampler towards nonzero regions of the grasp success probability function.
As we only focus on precision grasps, attracting grasps towards rims
(in the absence of any graspability information) is an effective heuristic,
since rims are very likely to allow precision grasps.
This is the only shape information required by our method.

The key contributions of this paper are:
\begin{itemize}
  \item A heuristic for calculating a pseudo measure for infeasible grasps to overcome
    the obstacle of applying MCMC for sampling gripper poses from a grasp density function
    where for infeasible grasps no quality measure can be calculated.
  \item The application of kernel adaptive MCMC for learning of precision grasps by active sampling.
  \item A transfer learning scheme for learning of precision grasps for
    novel objects by using suitable prior information by known gripper poses for
    similar objects, thus facilitating generalization of grasps.
\end{itemize}
We evaluate our proposed learning methods by a series of carefully designed experiments as
presented in Section~\ref{sec:6}. We conclude our work in Section~\ref{sec:7}.

\section{Related Work} \label{sec:2}

Traditionally, grasp learning methods rely on vision
for both (i) finding graspable parts of the object and (ii) evaluating
the learned grasps. Detry et al.~\cite{detry2011} learn grasp affordance densities
by (i) establishing a grasp affordance model for an object, and (ii) training this
model by autonomous exploration, e.g., sampling, where grasp affordance densities are
modeled by Kernel Density Estimation. Based on early visual cues an initial density is
created which is then trained and finally yields the empirical density of the object's
grasps. Detry et al.~\cite{detry2013} study transfer learning of
grasping strategies. Their method is based on learning object shape prototypes to
generalize grasping strategies among different objects. Kopicki et al.~\cite{kopicki2014}
propose to learn grasps by computing a gripper's shape as to a specific grasp relative to
an object's shape. Their method allows transfer of grasps by
matching the gripper's shape to shapes of novel objects. Fischinger et al.~\cite{fischinger2013}
investigate grasping objects from cluttered scenes on the grounds of a point cloud of the scene.
Using Symmetry Height Accumulated Features their system is trained by Support Vector
Machines yielding grasp classifiers which subsequently allow a robot to decide
on an optimal 6D gripper pose relative to its environment. The work of Kroemer
et al.~\cite{kroemer2010} combines the two notions of active learning and reactive control
into a grasp learning framework. Their system implements a hybrid architecture, where,
on the high-level side, a reinforcement learner (the active learner) determines grasps
and, on the low level, a reactive controller is responsible for action execution. Further,
vision is used to incorporate geographic scene information for optimal grasp learning.
Recently, Lenz et al.~\cite{lenz2013} applied deep learning for learning grasps
from an RGB-D view of a scene. Their method ultimately implements rectangle-based grasp
detection~\cite{jiang2011} using two neural networks, i.e., (i) for detecting potential
candidate rectangles on the object, and (ii) to extract top rectangles from these
candidate rectangles. The top rectangles then represent optimal grasps with respect to a gripper's
pose. Rodriguez et al.~\cite{rodriguez2011} suggest early abort and retry to reduce the
time to learn a grasp. In their work, grasp signatures are used to establish
probabilistic models to track the instantaneous probability of a grasp to succeed.
Given that the model suggests that a grasp may not succeed, it is aborted early and
retried using slightly modified parameters. Abort and retry are modeled as a Markov
chain which subsequently can be used to minimize the time for learning grasps.
Saxena et al.~\cite{saxena2011} introduce a vision-based grasping system
which learns grasping points for images of cluttered scenes. By supervised learning,
their system learns visual features for identifying a 3-D point and an orientation at
which to grasp the object. Given this information, a path planner then calculates the
optimal trajectory to reach the object and apply a grasp. Stulp et al.~\cite{stulp2011}
investigate learning grasps with a special emphasis on uncertainty. Their key
idea in learning optimal grasp poses is to sample actual object poses from a
distribution that represents the state estimation uncertainty. Dynamic
movement primitives necessary to reach the object are learned by reinforcement learning.

Our learning methods mainly differ in that they require only weak information from object models (such as
point clouds). Except for Detry et al.~\cite{detry2013} and Kopicki et al.\cite{kopicki2014}, most
other research does not address transfer learning. Our approach to transfer learning differs
in that we do not rely on precise shape information to generalize grasping strategies but instead
only on object poses.  The sole purpose of the object's rim points is
to keep the sampler from deterioration. Our method would still find grasps without them,
but it would take longer, as the sampler would degenerate to a purely random walk. In
terms of heuristic search strategies, the work of Stulp et al.~\cite{stulp2011} is related
to ours. Yet, contrary to our grasp learning method, Stulp et al.\ rely on an initial,
feasible grasp and preshape posture for their method to work. As Table~\ref{tab:res_learning}
shows, our grasp learning method, in contrast, also learns feasible grasps from random
initialization.
%%%%%%%%%%%%%%%%%%%%%%%%%%%%%%%%%%%%%%%%%%%%%%%%%%%%%%%%%%%%%%%%%%%%%%%%%%%%%%%%%%%%%%%%%%%%%%%%%%%%%%%%%%%%%%%%%%%%%%%%%%%%%%%%%%%%%%%%%%%%%%%%%%%%%%%
\section{Kernel-Adaptive MCMC} \label{sec:3}

MCMC Kameleon as proposed by Sejdinovic et al.~\cite{sejdinovic13} is an adaptive
MH sampler approximating highly non-linear target densities $\pi$. During
its burn-in phase, at each iteration it obtains a subsample
$\mathbf{z} = \left\{ z_{i} \right\}_{i=1}^{n}$ of the chain history
$\left\{ x_{i} \right\}_{i=0}^{t-1}$ for updating the proposal distribution
$q_{\mathbf{z}}(\cdot \mid x)$ by applying kernel PCA on $\mathbf{z}$,
resulting in a low-rank covariance operator $C_{\mathbf{z}}$.
Using $\nu^{2}C_{\mathbf{z}}$ as a covariance (where $\nu$ is a scaling parameter),
a Gaussian measure with mean $k(\cdot,y)$, i.e., $\mathcal{N}(f; k(\cdot,y),
\nu^{2}C_{\mathbf{z}})$, is defined. Samples $f$ from this measure are subsequently
used to obtain target proposals $x^{*}$.

MCMC Kameleon computes pre-images $x^{*} \in \mathcal{X}$ of $f$ by considering the
 non-convex optimization problem
\begin{equation}
  \argmin_{x\in\mathcal{X}}g(x),
\end{equation}
where
\begin{IEEEeqnarray}{rCl}
  g(x) &=& \left\Vert k\left(\cdot,x\right)-f\right\Vert _{\mathcal{H}_{k}}^{2} \\
  &=& k(x,x) - 2k(x,y) - 2 \sum_{i=1}^{n}\mathbf{\beta}_{i}\left [ k(x,z_{i})-\mu_{\mathbf{_z}}(x) \right ], \nonumber
\end{IEEEeqnarray}
$\mu_{\mathbf{z}} = \frac{1}{n}\sum_{i=1}^{n}k(\cdot,z_{i})$,
the empirical measure on $\mathbf{z}$, and $y \in \mathcal{X}$.
Then, by taking a single gradient descent step along the cost function $g(x)$
a new target proposal $x^{*}$ is given by
\begin{equation}
  x^{*} = y - \eta \nabla_{x}g(x)\rvert_{x=y} + \xi
  \label{eqn:MCMC.opt.step}
\end{equation}
where $\mathbf{\beta}$ is a vector of coefficients, $\eta$ the gradient step size,
and $\xi \sim \mathcal{N}(0,\gamma^{2}I)$ an additional isotropic exploration term after
the gradient.
 The complete MCMC Kameleon algorithm then is
\begin{itemize}
  \item at iteration $t+1$
  \begin{enumerate}
    \item obtain a subsample $\mathbf{z} = \left\{ z_{i} \right\}_{i=1}^{n}$ of the chain history $\left\{ x_{i} \right\}_{i=0}^{t-1}$,
    \item sample $x^{*} \sim q_{\mathbf{z}}(\cdot \mid x_{t}) = \mathcal{N}(x_{t},\gamma^{2}I + \nu^{2}M_{\mathbf{z},x_{t}}HM_{\mathbf{z},x_{t}}^{T})$,
    \item accept $x^{*}$ with MH acceptance probability $\alpha(x,y) = \min\begin{Bmatrix}
    1, \frac{\pi(y) q(x \mid y)}{\pi(x) q(y \mid x)}
  \end{Bmatrix}$,
  \end{enumerate}
\end{itemize}
where $M_{\mathbf{z},y}=2\eta\left[\nabla_{x}k(x,z_{1})|_{x=y},\ldots,\nabla_{x}k(x,z_{n})|_{x=y}\right]$
is the kernel gradient matrix obtained from the gradient of~\eqref{eq:cost}
at $y$, $\gamma$ is a noise parameter, and $H$ is an $n \times n$ centering matrix.

%%%%%%%%%%%%%%%%%%%%%%%%%%%%%%%%%%%%%%%%%%%%%%%%%%%%%%%%%%%%%%%%%%%%%%%%%%%%%%%%%%%%%%%%%%%%%%%%%%%%%%%%%%%%%%%%%%%%%%%%%%%%%%%%%%%%%%%%%%%%%%%%%%%%%%%
\section{Active Grasp Learning} \label{sec:4}

We represent a grasp $g$ as a 7D vector, i.e., $g = (x,y,z,q_{w},q_{x},q_{y},q_{z})^{T}$,
where $x$, $y$, $z$ denote the cartesian coordinates of a gripper and
$q_{w}$, $q_{x}$, $q_{y}$, $q_{z}$ its orientation in quaternion notation relative
to an object. To each grasp $g$ is associated a measure $\mu_{GWS}$ based on the Grasp Wrench
Space (GWS)~\cite{miller1999}, indicating its quality. This measurability
then allows us to define a target density $\pi(g), g \in \mathcal{X}$.

\subsection{Metropolis Criteria} \label{sec:4:1}

The GWS is only defined for a feasible grasp $g$; i.e., if a grasp $g$ cannot be
applied to an object, $\mu_{GWS}$ cannot be calculated.  Therefore, a heuristic is needed to calculate
approximate measures for samples $g$ that do not represent a feasible grasp. This stems from
MCMC's need for continuous probability densities to efficiently converge to the
global optimum by the MH acceptance criterion.

\subsubsection{Feasible Grasps}

In the case of a feasible grasp $g$, the probability measure is calculated
in terms of the GWS\@. Thus, $\pi(g) \propto \mu_{GWS}$,
and the acceptance probability of
the proposed sample $g$ is calculated by the MH acceptance criterion. The likelihood of
a new proposal $g^{*}$ conditioned by $g$ is given by the proposal distribution $q$'s
density for $g^{*}$, conditioned by $g$, i.e., $q(g^{*} \mid g)$. Conversely, the
likelihood of $g$ conditioned on $g^{*}$ is given by $q(g \mid g^{*})$.

\subsubsection{Infeasible Grasps}

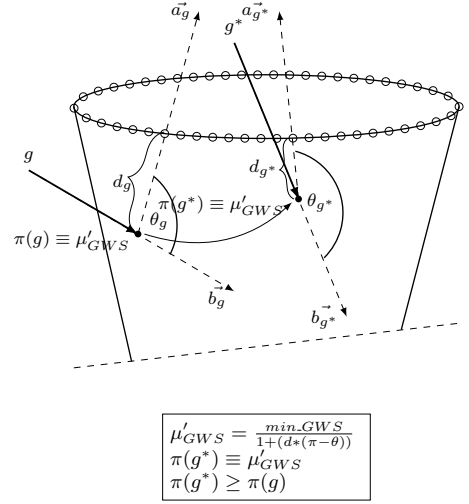
\begin{figure}[htb]
  \tikzset{paint/.style={ draw=#1!50!black, fill=#1!50 }, decorate with/.style={decorate,decoration={shape backgrounds,shape=#1,shape size=3.5pt}}}
  \centering
  \resizebox{.35\textwidth}{!}{
  \begin{tikzpicture}
    \tikzset{>=latex}
    %\draw[semithick] (0,0) arc (-180:0:2 and 0.25);
    %\draw[dashed,color=gray] (0,0) arc (180:0:2 and 0.25);
    %\draw[semithick] (0,0) -- (-1,6);
    %\draw[semithick] (4,0) -- (5,6);
    \draw[-,dashed] (-1,1.9) -- (5,2.6); %for cut off mug
    \draw[semithick] (-0.1,2) -- (-1,6);
    \draw[semithick] (4.1,2.5) -- (5,6);
    \draw[semithick] (2,6) ellipse (3 and 0.5);
    \draw[decorate with=circle] (2,6) ellipse (3 and 0.5);

    \draw[arrows=->, thick] (-1.7,5)--(0,4);
    \fill (0,4) circle[radius=1.5pt];
    \fill[black,font=\footnotesize] (0,3.9) node [left] {$\pi(g) \equiv \mu_{GWS}^{\prime}$};
    \fill[black,font=\footnotesize] (-1.7,5) node [above] {$g$};
    \draw[arrows=->, dashed] (-1.7,5)--(1.5,3.1);
    \fill[black,font=\footnotesize] (1.5, 3) node [left] {$\vec{b_{g}}$};
    \draw[arrows=->, dashed] (0,4)--(0.95,7.5);
    \fill[black,font=\footnotesize] (0.95,7.5) node [left] {$\vec{a_{g}}$};
    \draw[decorate,decoration={brace,amplitude=10pt}, xshift=-4pt,yshift=0pt](0.1,4) -- (0.55,5.6) node [black,midway,xshift=-12pt,font=\footnotesize] {$d_{g}$};
    \draw[semithick] (0.25,4.9) arc (50:-27:1 and 1);
    \fill[black,font=\footnotesize] (0.6,4.2) node [left] {$\theta_{g}$};

    \draw[arrows=->, thick] (1.5,7)--(2.5,4.55);
    \fill (2.5,4.55) circle[radius=1.5pt];
    \fill[black,font=\footnotesize] (2.4,4.45) node [left] {$\pi(g^{*}) \equiv \mu_{GWS}^{\prime}$};
    \fill[black,font=\footnotesize] (1.5,7) node [above] {$g^{*}$};
    \draw[arrows=->, dashed] (1.5,7)--(3.25,2.7);
    \fill[black,font=\footnotesize] (3.25,2.7) node [left] {$\vec{b_{g^{*}}}$};
    \draw[arrows=->, dashed] (2.5,4.55)--(2.2,7.5);
    \fill[black,font=\footnotesize] (2.2,7.5) node [left] {$\vec{a_{g^{*}}}$};
    \draw[decorate,decoration={brace,amplitude=5pt}, xshift=-2pt,yshift=0pt](2.5,4.55) -- (2.4,5.5) node [black,midway,xshift=-12pt,font=\footnotesize] {$d_{g^{*}}$};
    \draw[semithick] (2.45,5.3) arc (75:-45:1 and 1);
    \fill[black,font=\footnotesize] (3.2,4.5) node [left] {$\theta_{g^{*}}$};

    \path[->,bend right] (0.1,4) edge (2.4,4.5);
    \node[draw,font=\small,text width=3cm] at (2,0.5) {$\mu_{GWS}^{\prime} = \frac{min\_GWS}{1 + (d * (\pi - \theta))}$\\$\pi(g^{*}) \equiv \mu_{GWS}^{\prime}$\\$\pi(g^{*}) \geq \pi(g)$};
  \end{tikzpicture}
  }
  \caption{Illustration of our heuristic used to calculate $\mu_{GWS}^{\prime}$.
    Here, $\pi$ denotes the mathematical constant and $\pi(\cdot)$
    a probability density.}
  \label{fig:heuristic}
\end{figure}

In the case of an infeasible grasp $g$, $\mu_{GWS}$ is zero, implying that
the MH acceptance probability would always be zero. This eventually robs the algorithm
of any clue whether the current search direction is promising or not. To overcome this
problem, we apply
a heuristic to learn whether an infeasible proposal $g^{*}$ points into a direction where a
feasible grasp could be found or not. Figure~\ref{fig:heuristic} illustrates
how we calculate this heuristic quality measure $\mu_{GWS}^{\prime}$.

The idea underlying the heuristic quality measure $\mu_{GWS}^{\prime}$ is based
(i) on the angle $\theta_{g}$ between the gripper's pose and the vector between the
gripper direction and a closest rim point as shown in Figure~\ref{fig:heuristic}
by the vector $\vec{a}_{g}$, and (ii) the distance of the gripper position and a
closest rim point indicated by $d_{g}$.
Setting the constant $min\_GWS$ to $0.01$, i.e., the lowest tolerable quality measure for a
grasp $g$, and using $\mu_{GWS}^{\prime}$ as defined in Figure~\ref{fig:heuristic},
we move the search for grasps $g^{*}$ towards the rims of an object.

To detect rim points of objects by their point clouds, first, for
each point $o$, we construct a spherical neighborhood $N$ around $o$. Next, we
connect $o$ to each of its neighboring points $i \in N$ to obtain vectors $\vec{p}_i$.
Then, if $o$ is a non-rim point and $N$ is flat, $\left \| \sum_{i \in N} \vec{p}_{i} \right \|^{2} \approx 0$.
However, if $o$ is a rim point, then $\left \| \sum_{i \in N} \vec{p}_{i} \right \|^{2} > \zeta$,
where $\zeta$ is a threshold that is tuned based on the density and noise of the point
clouds. For all other non-rim points, $0 < \left \| \sum_{i \in N} \vec{p}_{i} \right \|^{2} \leq \zeta$.
Figures~\ref{fig:object_set_grasp_learning} and~\ref{fig:object_set_transfer_learning}
show the results of the rim detection applied to the objects used in our experiments (Section~\ref{sec:6}).
%\begin{figure}[t]
%  \centering
%    \includegraphics[width=.45\textwidth]{margin_extraction}
%    \caption{Rim point detection (best viewed in color).}
%    \label{fig:rim_extraction}
%\end{figure}

Observe that both quality measures, i.e., $\mu_{GWS}$ and $\mu^{\prime}_{GWS}$, are valid
density functions in that (i) their values are always greater or equal zero and (ii)
by introducing some normalization constant $Z$, i.e., $Z = \sum_{i=1}^{n} \mu_{GWS}^{i}$
(where $n$ is the number of known, feasible grasps for an object), applied to
$\mu_{GWS}$, i.e., $\frac{1}{Z}\mu_{GWS}$, we have that $\int\pi(g)\mathrm{d}g = 1$ (similarly,
this also holds for $\mu_{GWS}^{\prime}$).

\subsection{Simulated Annealing} \label{sec:4:2}
Using a plain Metropolis-Hastings (MH) acceptance criterion $\alpha(g^{*},g) = \min
  \begin{Bmatrix}
    1, \frac{\pi(g^{*}) q(g^{*} \mid g)}{\pi(g) q(g \mid g^{*})}
  \end{Bmatrix}$,
MCMC Kameleon considers the whole search space $\mathcal{X}$,
i.e., the Markov chain will likely visit bad samples as well as good ones.
In learning to grasp objects, however, we are only interested in
good samples, i.e., feasible grasps. To tackle this problem we apply simulated
annealing (SA)~\cite{van1987}. The idea is to equip the sampler with an
initial temperature $T>0$ which decreases over the sampling process
with the effect of gradually decreasing the probability of accepting poor samples
while exploring the state space $\mathcal{X}$. Consequently, while traversing the
Markov chain the sampler more likely moves in regions most likely containing
the global optima. We thus extend the plain MH acceptance criterion
by raising it to the power of $T$, where $T$ is the current system temperature, i.e.,
\begin{IEEEeqnarray}{rCl}
  \label{eq:mh:accept_sa}
  \alpha(x,y) &=& \min\begin{Bmatrix}
    1, \frac{\pi(g^{*}) q(g \mid g^{*})}{\pi(g) q(g^{*} \mid g)}
  \end{Bmatrix}^{T}
  \label{eq:temp} \\
  T &=& \max\begin{Bmatrix}
    T_{N}, \frac{T_{N}}{T_{0}}^{\frac{j}{N}}
  \end{Bmatrix}
\end{IEEEeqnarray}
where $T_{0}$ is the initial temperature, $T_{N}$ the final temperature, $N$
the number of iterations, and $j$ the current iteration. Using~\eqref{eq:mh:accept_sa}
we slowly decrease the acceptance probability of bad, i.e., of low quality,
grasp proposals $g^{*}$. Equation~\eqref{eq:temp} slowly decreases the temperature $T$
over time.

\subsection{Complete Learning Method} \label{sec:4:3}

We use a Gaussian proposal for both position \emph{and} orientation to
capture the relation between gripper position and orientation relative
to an object. Further, as a kernel $k$ for MCMC Kameleon we use a
Gaussian kernel. Although Gaussian proposals and kernels are not
rigorously applicable in quaternion space, this choice allows us to
easily model the dependencies between gripper positions and
orientations, which is crucial for our method to perform well.

Our complete learning algorithm then is
\begin{itemize}
  \item at iteration $t+1$
  \begin{enumerate}
    \item obtain a subsample $\mathbf{z} = \left\{ z_{i} \right\}_{i=1}^{n}$ of the chain history $\left\{ g_{i} \right\}_{i=0}^{t-1}$,
    \item sample $g^{*} \sim q_{\mathbf{z}}(\cdot \mid g_{t}) = \mathcal{N}(g_{t},\gamma^{2}I + \nu^{2}M_{\mathbf{z},g_{t}}HM_{\mathbf{z},g_{t}}^{T})$,
    \item calculate $\pi(g^{*})$ using either $\mu_{GWS}$ in the case of a feasible grasp or $\mu_{GWS}^{\prime}$ otherwise
    \item accept $g^{*}$ with MH acceptance probability \eqref{eq:mh:accept_sa}.
  \end{enumerate}
\end{itemize}
The chain history $\left\{ g_{i} \right\}_{i=0}^{t-1}$ is initialized by samples as
retrieved\footnote{These samples amount to the ``rough sketch'' (Sec.~\ref{sec:sketch}) of the shape
of the grasps' success probability.} from a random walk (RW) MCMC sampler using a
Gaussian proposal for the position and a von Mises-Fisher proposal for the orientation,
i.e.,
\begin{equation*}
  \begin{split}
    &g^{*}_{pos} = \mathcal{N}(g_{pos}^{t}, \Sigma)\\
    &g^{*}_{ori} = \mathcal{C}_{4}(\kappa)\exp(\kappa g_{ori}^{t\;\ \top} \mathbf{x}),
  \end{split}
\end{equation*}
where $\kappa$ is the concentration parameter and $\mathbf{x}$ a p-dimensional
unit direction vector. We use the same probability measures as defined
for MCMC Kameleon.

%%%%%%%%%%%%%%%%%%%%%%%%%%%%%%%%%%%%%%%%%%%%%%%%%%%%%%%%%%%%%%%%%%%%%%%%%%%%%%%%%%%%%%%%%%%%%%%%%%%%%%%%%%%%%%%%%%%%%%%%%%%%%%%%%%%%%%%%%%%%%%%%%%%%%%%
\section{Transfer Learning} \label{sec:5}

Humans apply transfer learning on a daily basis by reusing acquired knowledge
and applying it to solve a problem similar to the one the knowledge originally was
learned for. Similarly, here we can reuse a chain
history $\left\{ g_{i} \right\}_{i=0}^{t-1}$ that was learned for some object
as a rough sketch from which to learn grasps $g$ for similar objects by properly biasing,
i.e., initializing, MCMC Kameleon with such a chain history. This is feasible thanks to
the burn-in phase of MCMC Kameleon where it learns an approximation of the target
density $\pi$. Given that two objects $a$ and $b$ are of similar shape and size
(e.g., a plate and a soup plate), the algorithm from Section~\ref{sec:4:3} can be
used for transfer learning of grasping a novel object $b$ given a chain history for object $a$.
Since producing an initial chain history for a novel object is
expensive, transfer learning by avoiding the burn-in phase can result
in substantial savings.

%%%%%%%%%%%%%%%%%%%%%%%%%%%%%%%%%%%%%%%%%%%%%%%%%%%%%%%%%%%%%%%%%%%%%%%%%%%%%%%%%%%%%%%%%%%%%%%%%%%%%%%%%%%%%%%%%%%%%%%%%%%%%%%%%%%%%%%%%%%%%%%%%%%%%%%
\section{Experiments} \label{sec:6}

In the following, we discuss (i) our results on learning grasps for a
given object (Section~\ref{sec:6:2}), and (ii) our results on transfer
learning of learning grasps for previously unseen objects
(Section~\ref{sec:6:3}). Both our learning algorithms were implemented in Python.
As a simulation environment for executing a sampled grasp $g$ and subsequently
calculating the GWS we used RobWork~\cite{ellekilde2010}. Inside RobWork,
all grasps were executed with a Schunk SDH gripper.

%\begin{figure}
%  \centering
%  \includegraphics[width=.4\columnwidth]{pan_scene}
%  \caption{Screen shot from RobWork when simulating a successful grasp with Schunk SDH gripper for the pan from Figure~\ref{fig:object_set_grasp_learning}.}
%  \label{fig:simulation}
%\end{figure}

\subsection{Grasp Learning} \label{sec:6:2}

\begin{figure}
  \centering
  \resizebox{\columnwidth}{!}{%
    \includegraphics{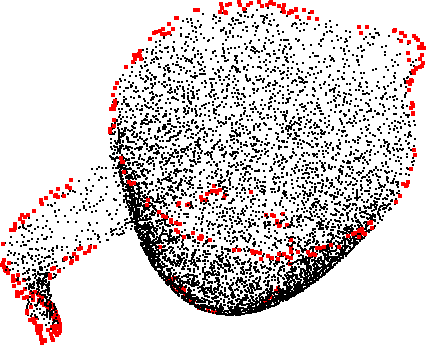}\hspace{1cm}%
    \includegraphics{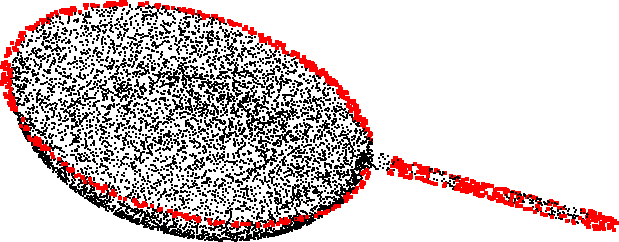}%
    \includegraphics{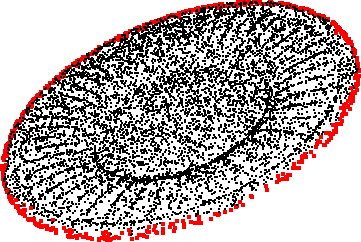}}
  \caption{Object set used for grasp learning with object rims depicted
    in red (best viewed in color).}
  \label{fig:object_set_grasp_learning}
\end{figure}

Figure~\ref{fig:object_set_grasp_learning} depicts the object set we used for
learning grasps of given objects. The experiments were done in two steps:
First, we ran a RW MCMC sampler for 5000 iterations to produce initial chain histories
$\left\{ g_{i} \right\}_{i=0}^{t-1}$. Secondly, we used these chain histories (as ``rough
sketches'') to initialize MCMC Kameleon for learning an approximation of the target
density $\pi$ during burn-in and subsequent grasp sampling (after burn-in).
Apart from the simulations just mentioned, we further did a series of experiments where
MCMC Kameleon was initialized with a randomly generated chain, without using prior knowledge.

Table~\ref{tab:params} shows our parametrization of MCMC Kameleon used during our
experiments\footnote{Our RW MCMC sampler offers two parameters, viz.~$\kappa$ and
$\Sigma$ which were set to $5$ and $0.01I_{3}$ for all experiments.}. Apart from
the value of the scaling parameter $\nu$ which was chosen following the
suggestion by Sejdinovic et al.~\cite{sejdinovic13}, the parameters
were tuned by a series of preliminary experiments.
\begin{table}
  \renewcommand{\arraystretch}{2}
  \centering
  \caption{Parameters for MCMC Kameleon used during our experiments.}
  \begin{tabular}{@{}ccccccc@{}}
    Iterations (N) & $\gamma$ & Subsample size& $\nu$ &
    Burn-in    & $T_{0}$ & $T_{N}$ \\\hline
    5000       & 0.0001   & 200           & $\frac{2.38}{\sqrt{6}}$ &
    \begin{minipage}[c]{3em} \centering 1000\\2000 \end{minipage}
    & 1.0     & 0.05
  \end{tabular}
  \label{tab:params}
\end{table}

For each of the objects from Figure~\ref{fig:object_set_grasp_learning} we
thus conducted one initial random walk for establishing a chain history
$\left\{ g_{i} \right\}_{i=0}^{t-1}$, and then four runs using MCMC Kameleon
for learning grasps with either a randomly initialized chain (no prior information),
or using chain histories from the random walks (using prior information), and
differing burn-in durations (Table~\ref{tab:params}). Table~\ref{tab:res_learning} shows
our results. Figure~\ref{fig:results} (top row) shows sampled gripper poses for
the objects from Figure~\ref{fig:object_set_grasp_learning}.
\begin{table}
  \renewcommand{\arraystretch}{1.2}
  \centering
  \caption{Experimental results: number of sampled gripper poses yielding
    feasible grasps during grasp learning. The numbers in square brackets in the
    leftmost column denote the burn-in duration
    and the use of prior information (p) or not (np).}
  \begin{tabular}{l | C{.8cm} C{.8cm} C{.8cm}}
                                     & Pitcher & Pan & Plate \\\hline
    RW MCMC                          & 12      & 3   & 20 \\
    MCMC Kameleon [1000,np] & 12      & 10  & 35 \\
    MCMC Kameleon [1000,p]  & 2       & 34  & 20 \\
    MCMC Kameleon [2000,np] & 16      & 54  & 95 \\
    MCMC Kameleon [2000,p]  & 14      & 115 & 81
  \end{tabular}
  \label{tab:res_learning}
\end{table}

As is evident from Table~\ref{tab:res_learning}, the complexity of an object's
shape clearly drives the success of learning. Generally, the algorithms we used
(RW MCMC and MCMC Kameleon) learned more gripper poses yielding feasible grasps for
the pan and the plate than for the pitcher (except for the RW MCMC and one run of
MCMC Kameleon). This is due to the geometry of the pitcher, which is more complex than the
pan or the plate. Table~\ref{tab:res_learning} also shows that using prior
information generally fosters learning of grasps. MCMC Kameleon
generally outperformed RW MCMC thanks to burn-in where it learns an approximation
of $\pi(g)$.

Another interesting aspect of MCMC Kameleon that is evident from Table~\ref{tab:res_learning}
is the choice of the burn-in duration. Using a longer burn-in results in better learning fo $\pi$,
 which generally results in a substantially larger number of grasps found.
Here, we kept the total number of iterations fixed to 5000: If the burn-in duration lasted for
1000 iterations, then MCMC Kameleon had a budget of 4000 iterations for sampling grasps; if
the burn-in lasted for 2000 iterations, it had a budget of 3000 iterations.

\subsection{Transfer Learning} \label{sec:6:3}

\begin{figure}
  \centering
  \resizebox{\columnwidth}{!}{%
    \includegraphics{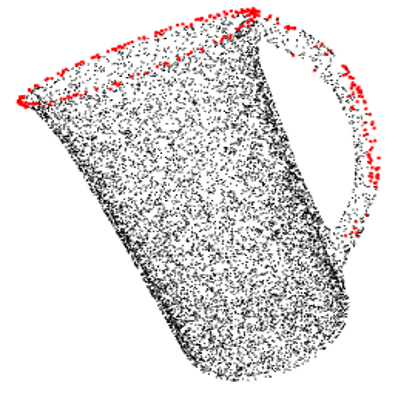}\hspace{1cm}%
    \includegraphics{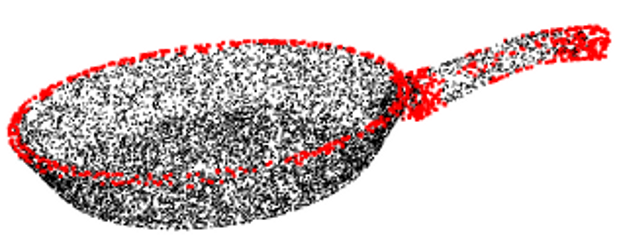}%
    \includegraphics{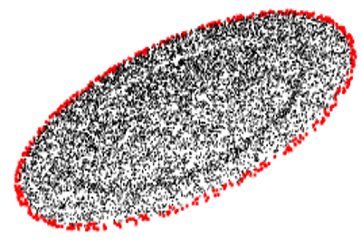}}\vspace{.5cm}
  \resizebox{\columnwidth}{!}{%
    \includegraphics{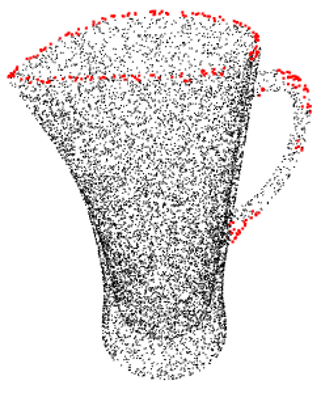}\hspace{1cm}%
    \includegraphics{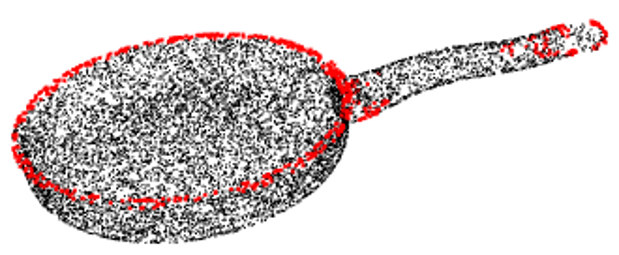}%
    \includegraphics{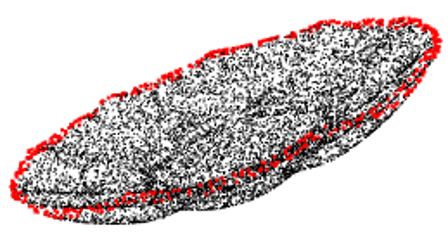}}
  \caption{Object set used for transfer learning with object rims depicted
    in red (best viewed in color).}
  \label{fig:object_set_transfer_learning}
\end{figure}

For learning of grasps for novel (as of yet unseen) objects we consider two
methods for biasing MCMC Kameleon with prior information (representing a rough sketch
of the target grasp density $\pi(g)$). First, we reuse complete
chain histories $\left\{ g_{i} \right\}_{i=0}^{t-1}$ as generated by MCMC Kameleon for
similar objects. Secondly, we investigate using subsamples $\mathbf{z}$ as computed by
MCMC Kameleon at the last iteration of the burn-in when sampling for a similar object.
In the latter case we skip the burn-in.  This amounts to a precast
covariance operator $C_{\mathbf{z}}$, preventing MCMC Kameleon from learn a
better approximation of $\pi(g)$, limiting it to just using what it is provided with.
Figure~\ref{fig:object_set_transfer_learning} shows our object set for this purpose.

We ran MCMC Kameleon with the same parameters as discussed in Section~\ref{sec:6:2} (Table~\ref{tab:params}),
except for the number of iterations. In the case of using a subsample $\mathbf{z}$ for transfer
learning we reduced the default number of 5000 iterations, i.e., the number of iterations was set to
3000 and 4000 respectively, depending on the burn-in duration of the run that produced $\mathbf{z}$.
Both chain histories and subsamples for transfer learning are a result of the experiments from
Section~\ref{sec:6:2}.

\begin{figure}
  \centering
    \resizebox{\columnwidth}{!}{%
    \includegraphics{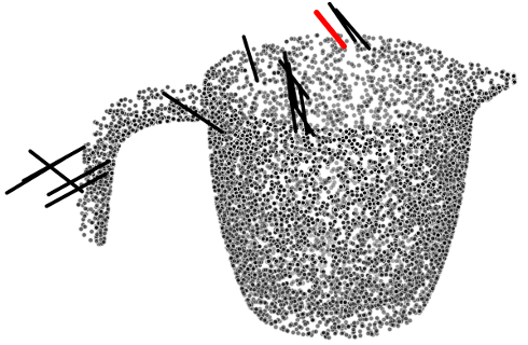}%
    \includegraphics{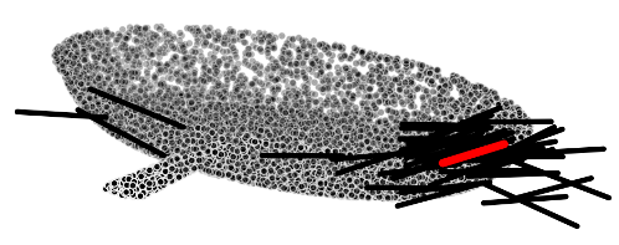}%
    \includegraphics{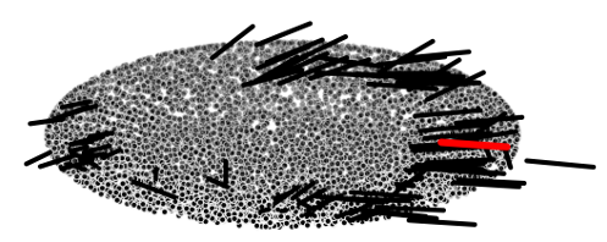}}
    \resizebox{\columnwidth}{!}{%
    \includegraphics{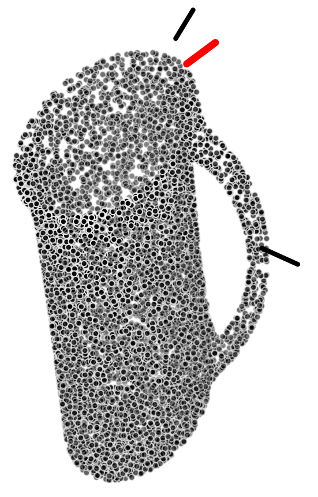}%
    \includegraphics{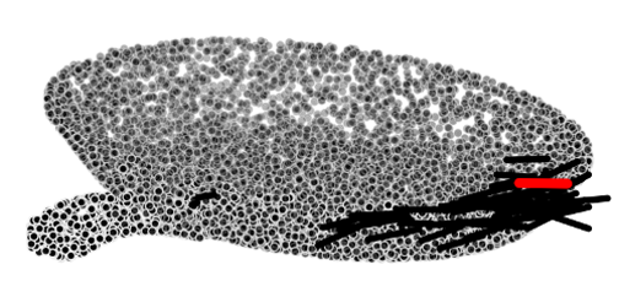}%
    \includegraphics{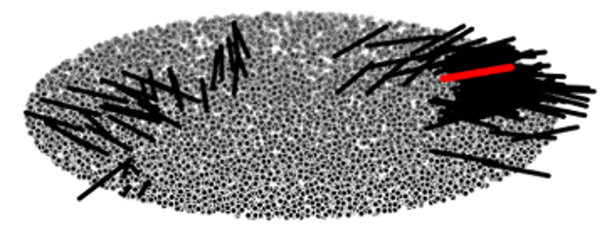}}
    \resizebox{\columnwidth}{!}{%
    \includegraphics{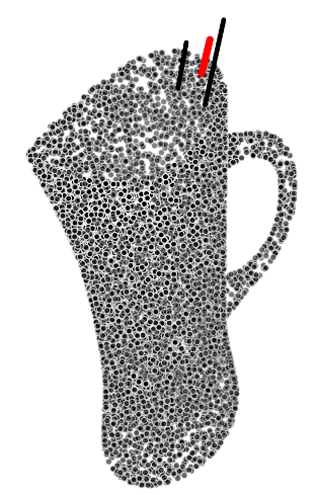}%
    \includegraphics{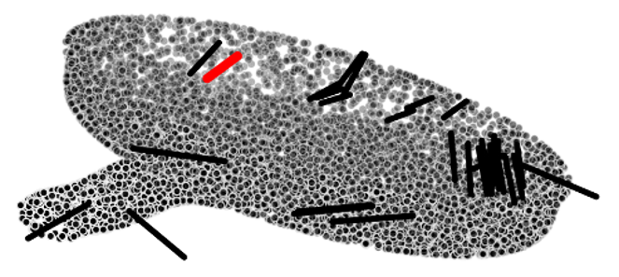}%
    \includegraphics{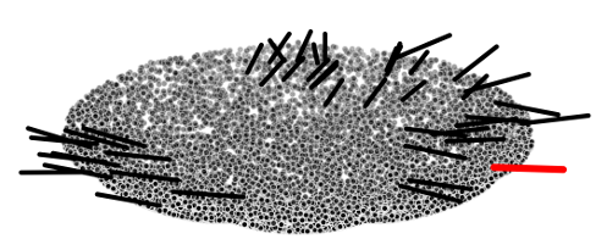}}
  \caption{Results for learning grasps for the objects from Figure~\ref{fig:object_set_grasp_learning}
    (top row) and for transfer learning of grasps for corresponding
    objects from Figure~\ref{fig:object_set_transfer_learning} (middle and bottom rows) with a burn-in duration of 1000. Red
    pointers indicate optimal grasps (best viewed in color). Observe that grasps are rather
    unevenly distributed; this results from both running MCMC Kameleon for only 5000 iterations
    and the use of SA which at some point locks the sampler to a mode of $\pi$.}
  \label{fig:results}
\end{figure}

Hence, for each object from Figure~\ref{fig:object_set_transfer_learning} we did (i) two runs with
MCMC Kameleon initialized by a chain history as established during an earlier
burn-in phase and a randomly selected starting point from that chain, and (ii) two runs with MCMC
Kameleon initialized with a kernel $\mathcal{K}$ where the starting point is chosen to be the
last element of the kernel $\mathcal{K}$. Observe that in the case of transfer learning, the objects
from Figure~\ref{fig:object_set_transfer_learning} first were aligned to a canonical pose to be
in line with the objects from Figure~\ref{fig:object_set_grasp_learning}. This is necessary as
chain histories and kernels $\mathcal{K}$ are established relative to an object's
orientation and size. Table~\ref{tab:res_tlearning} shows our results. Figure~\ref{fig:results}
(middle and bottom rows) shows sampled gripper poses for the objects from
Figure~\ref{fig:object_set_transfer_learning}.
\begin{table}
  \renewcommand{\arraystretch}{1.2}
  \centering
  \caption{Experimental results: number of sampled gripper poses yielding
    feasible grasps, for transfer learning of grasps. The second and fourth row correspond
    to runs with 4000 iterations, the sixth and eighth to runs with 3000 iterations. The
    arrangement of results corresponds to Figure~\ref{fig:object_set_transfer_learning}. The
    numbers in square brackets in the leftmost column denote the burn-in duration and the usage of a chain~(c) or a subsample ($\mathbf{z}$) to initialize MCMC
    Kameleon.}
  \begin{tabular}{l | C{.8cm} C{.8cm} C{.8cm}}
                                     & Pitcher & Pan & Plate \\\hline
    MCMC Kameleon [1000,c] & 3 & 53 & 117 \\
    MCMC Kameleon [\phantom{000}0,$\mathbf{z}$]    & 0 & 0  & 0 \\
    MCMC Kameleon [2000,c] & 2 & 19  & 7 \\
    MCMC Kameleon [\phantom{000}0,$\mathbf{z}$]    & 4 & 68 & 0 \\\hline
    MCMC Kameleon [1000,c] & 3 & 29 & 45 \\
    MCMC Kameleon [\phantom{000}0,$\mathbf{z}$]    & 0 & 0  & 0 \\
    MCMC Kameleon [2000,c] & 1 & 6  & 62 \\
    MCMC Kameleon [\phantom{000}0,$\mathbf{z}$]    & 0 & 41 & 0 \\
  \end{tabular}
  \label{tab:res_tlearning}
\end{table}

What first leaps to the eye in Table~\ref{tab:res_tlearning} are the rather poor
results we achieved for the pitcher from Figure~\ref{fig:object_set_transfer_learning}.
It turns out that this is due to object misalignment, i.e., the pitchers from
Figure~\ref{fig:object_set_transfer_learning} were not properly aligned with the pitcher from
Figure~\ref{fig:object_set_grasp_learning}. Apart from that, the similarity in dimensions also
plays a key role for transfer learning. Clearly, the two pitchers from Figure~\ref{fig:object_set_transfer_learning}
are taller than the one from Figure~\ref{fig:object_set_grasp_learning}, rendering learning with
MCMC Kameleon difficult as the initial rough sketch of the shape of the object's grasp density does
not fit well. Obviously, misalignment of objects further exacerbates this mismatch.

Our results from Table~\ref{tab:res_tlearning} show that if MCMC Kameleon is initialized with a
subsample $\mathbf{z}$ of an earlier run it may drastically fail. We conclude that this is because
a subsample $\mathbf{z}$ is already too specific a sketch of an object's grasp density, i.e., it does
not provide the necessary diversity to \enquote{recognize} similar objects. Clearly, by again
permitting MCMC Kameleon to employ a burn-in phase, such a subsample $\mathbf{z}$ could be adapted
to an object. In contrast, our results in Table~\ref{tab:res_tlearning} demonstrate that initializing MCMC Kameleon
by reusing a chain history $\left\{ g_{i} \right\}_{i=0}^{t-1}$ is very effective. We thus conjecture
that reusing a chain history, i.e., prior experience, together with a burn-in phase for transfer
learning results in a good number of feasible grasps for previously unseen objects. Moreover, reusing
such chain histories for transfer learning successfully eradicates the need for performing random walks
to obtain a rough sketch of the object, thus reducing the computational burden incurred by our learning methods.

To sum up, we can state that the learning methods introduced in this paper
have proven successful both in learning of grasps for given
objects (Section~\ref{sec:4}) and in learning of grasps for previously
unseen objects, i.e., applying transfer learning for learning of
grasps (Section~\ref{sec:5}).

%%%%%%%%%%%%%%%%%%%%%%%%%%%%%%%%%%%%%%%%%%%%%%%%%%%%%%%%%%%%%%%%%%%%%%%%%%%%%%%%%%%%%%%%%%%%%%%%%%%%%%%%%%%%%%%%%%%%%%%%%%%%%%%%%%%%%%%%%%%%%%%%%%%%%%%
\section{Conclusion} \label{sec:7}

In this paper we have introduced (i) active learning of grasps for given objects
and (ii) transfer learning for learning grasps for novel objects.
Both our learning methods build upon MCMC Kameleon, an adaptive MH sampler that
learns an approximation of a target density $\pi$ during its burn-in phase.
\balance

The experimental evaluation shows that the application of an adaptive MH sampler,
e.g., MCMC Kameleon, is promising for grasp learning tasks as discussed in this paper.
Our results show that our learning methods allow learning grasps from no knowledge
at all, as Table~\ref{tab:res_learning} shows, when MCMC Kameleon is initialized
with a random chain, i.e., it uses no prior information. Table~\ref{tab:res_learning} however
also shows that our learning methods can be easily boosted if MCMC Kameleon is initialized
with suitable prior experience, i.e., a rough sketch of the shape of the grasps' probability
density associated with an object. We successfully exploit this aspect of biased
initialization in transfer learning (Table~\ref{tab:res_tlearning}) of grasps
for novel objects.

%%%%%%%%%%%%%%%%%%%%%%%%%%%%%%%%%%%%%%%%%%%%%%%%%%%%%%%%%%%%%%%%%%%%%%%%%%%%%%%%%%%%%%%%%%%%%%%%%%%%%%%%%%%%%%%%%%%%%%%%%%%%%%%%%%%%%%%%%%%%%%%%%%%%%%%

%\addtolength{\textheight}{-12cm}

\bibliographystyle{IEEEtran}
\bibliography{references}

\end{document}